\documentclass[sn-mathphys-ay]{sn-jnl}

\usepackage{lscape}
 
\usepackage{times}
\usepackage{soul}
\usepackage{url}
\usepackage[utf8]{inputenc}
\usepackage[small]{caption}
\usepackage{subcaption}
\usepackage{graphicx}
\usepackage{amsmath,amssymb,amsfonts}
\usepackage{amsthm}
\usepackage{booktabs}
\usepackage{algorithm}
\usepackage{algorithmic}
\usepackage{amsmath,amssymb,amsfonts}
\usepackage{textcomp}
\usepackage{xcolor}
\usepackage{multirow}
\usepackage{enumerate}
\usepackage{arydshln}

\begin{document}
\title{A Survey on the Real Power of ChatGPT}

\author*[1]{\fnm{Ming} \sur{Liu}}\email{m.liu@deakin.edu.au}

\author[2]{\fnm{Ran} \sur{Liu}}\email{liuran@iie.ac.cn}

\author[1]{\fnm{Ye} \sur{Zhu}}\email{ye.zhu@ieee.org}

\author[3]{\fnm{Hua} \sur{Wang}}\email{hua.wang@vu.edu.au}

\author[4]{\fnm{Youyang} \sur{Qu}}\email{youyang.qu@data61.csiro.au}

\author[5]{\fnm{Rongsheng} \sur{Li}}\email{dasheng@hrbeeu.edu.cn}

\author[6]{\fnm{Yongpan} \sur{Sheng}}\email{shengyp2011@gmail.com}

\author[7]{\fnm{Wray} \sur{Buntine}}\email{wray.b@vinuni.edu.vn}

\affil*[1]{\orgname{Deakin University}, \orgaddress{ \postcode{3125}, \state{Victoria}, \country{Australia}}}

\affil[2]{\orgname{Chinese Academy of Sciences}, \orgaddress{\postcode{100864}, \state{Beijing}, \country{China}}}

\affil[3]{\orgname{Victoria University}, \orgaddress{ \postcode{8001}, \state{Victoria}, \country{Australia}}}

\affil[4]{\orgname{Data61}, \orgaddress{ \postcode{3168}, \state{Victoria}, \country{Australia}}}

\affil[5]{\orgname{Harbin Engineering University}, \orgaddress{ \postcode{150001}, \state{Harbin}, \country{China}}}

\affil[6]{\orgname{Southweat University}, \orgaddress{ \postcode{400715}, \state{Chongqing}, \country{China}}}

\affil[7]{\orgname{VinUnivesity}, \orgaddress{ \postcode{9015}, \state{Hanoi}, \country{Vietnam}}}
 



\abstract{
ChatGPT has changed the AI community and an active research line is the performance evaluation of ChatGPT. A key challenge for the evaluation is that ChatGPT is still closed-source and traditional benchmark datasets may have been used by ChatGPT as the training data. In this paper, (i) we survey recent studies which uncover the real performance levels of ChatGPT in seven categories of NLP tasks, (ii) review the social implications and safety issues of ChatGPT, and (iii) emphasize key challenges and opportunities for its evaluation. We hope our survey can shed some light on its blackbox manner, so that researchers are not misleaded by its surface generation.}

\maketitle

\section{Introduction}
It has been more than a year since OpenAI's release of ChatGPT (Chat Generative Pre-trained Transformer) \footnote{\url{https://openai.com/chatgpt}}. According to the latest statistic \footnote{\url{https://explodingtopics.com/blog/chatgpt-users}} in Jan 2024, ChatGPT currently has over 180.5 million monthly users and \url{openai.com} gets approximately 1.5 billion visits per month. ChatGPT is built upon either GPT-3.5 or GPT-4, both of which are members of OpenAI's proprietary series of generative pre-trained transformer models, based on the transformer architecture and is fine-tuned for conversational applications using a combination of supervised learning and reinforcement learning.   Although its success has changed the ecosystem of the AI and NLP community, this large language model (system) is still a black box and researchers do not have much knowledge about the training details. Traditional evaluation approaches rely on training and test split with some benchmark datasets, these methods may not be available as these data are very likely to be used in the training phase of ChatGPT. Therefore, new leaderboards such as the chatbot-arena-leaderboard \footnote{\url{https://huggingface.co/spaces/lmsys/chatbot-arena-leaderboard}} incorporate both automatic and human voting. However, these leaderborads only show the general ability of ChatGPT and do not tell the exact performance on specific NLP tasks. 

In this paper, we review the more recent papers in the ML and NLP community for ChatGPT evaluation. Specifically, we survey the following three aspects: i) ChatGPT's performance on seven main categories of NLP tasks. ii) the social implication and safety issues of ChatGPT. (iii) the performance of ChatGPT over time. We also highlight key challenges and opportunities of ChatGPT's evaluation. Our key findings are: First, ChatGPT's performance tends to be good in the zero and few shot settings, but still under-perform the fine tuned models. Second, the generalization ability of ChatGPT is limited when it is evaluted on newly collected data. Third, most evaluation works utilize prompt engineering, which rely on human heuristics and cannot guarantee reproducibility. Last but not least, the performance of ChatGPT degrades with time. 

\section{Large Language Modeling}  \label{sec:define}

\subsection{Modeling Paradigm}
In the statistical age, n-gram language modelling calculate n gram statistics and use rules or Markov models for language modelling. These techniques often lose key text information (e.g. word order) and cannot handle out-of-vocabulary words. Therefore, smoothing techniques are widely used to avoid zero probability to unseen or infrequent n-grams. In the neural age, the idea of distributed representation basically learn high dimensional semantic embedding for words with deep neural networks. Self-supervised learning is often used to design the learning objectives, e.g, next token prediction, neighbour sentence classification. A few big milestones for neural language modelling include word2vec, BERT and the GPT models. 
In terms of the neural architectures for language modelling: there are encoder-only (e.g. BERT), encoder-decoder based (e.g. BART, T5) and decoder-only (e.g. GPT3). Recent trend in industry has demonstrated the appealing ability of decoder-only architecture.  
\subsection{Open Models}
Many open-source large language models have been competitive against the commercial alternatives, especially once fine-tuned and optimized. A key strength for open-source models is that they do not require individuals or enterprises to send their data to third-party remote servers, thus protecting user privacy. A few open-source pre-trained large language models include LLaMA by Meta \footnote{\url{https://ai.meta.com/llama/}}, Mistral 7B by Mistral \footnote{\url{https://mistral.ai/news/announcing-mistral-7b/}}, Falcon LLM by TII \footnote{\url{https://falconllm.tii.ae/}}, GPT2 by OpenAI \footnote{\url{https://openai.com/research/gpt-2-1-5b-release}}, GPT-J by EleutherAI \footnote{\url{https://www.eleuther.ai/artifacts/gpt-j}}, MPT by MosaicML \footnote{\url{https://www.mosaicml.com/mpt}}, BLOOM by BigScience \footnote{\url{https://huggingface.co/bigscience/bloom}}, ChatGLM 6B by Zhipu \footnote{\url{https://github.com/THUDM/ChatGLM-6B}}. LlaMA 2 is now widely used in the research community. It is trained on 2 trillion tokens and have double the context length than LlaMA 1. Its fine-tuned models have been trained on over 1 million human annotations. 

\subsection{Closed Models} 
To the date of our writing, ChatGPT and GPT-4 are the two widely accepted commercialized systems. The general training process of ChatGPT includes self-supervised training, reward model learning and reinforcement learning, but it is still unknown how much training data and human annotation is used in training ChatGPT. It is not confirmed from OpenAI that GPT-4 is a mixture-of-experts system in which eight 220B experts were trained with different data and task distributions. GLaM \footnote{\url{https://gpt3demo.com/apps/google-glam}} introduced the sparsely activated mixture-of-experts architecture to scale the model capacity while also incurring substantially less training cost compared to dense variants. As a result, The largest GLaM has 1.2 trillion parameters, which is approximately 7x larger than GPT-3.

\section{ChatGPT Performance Evaluation} \label{sec:performance_evaluation}
In this section, we review recent works which directly use ChatGPT for specific NLP tasks. The tasks include but are not limited to: classification, text generation, sequence labelling, information retrieval, parsing, reasoning, multilingualism, and other mixed tasks. Table 1 shows ChatGPT's performance on some of the NLP tasks, we will illustrate each of these tasks in the following sections.
\subsection{Classification}
Traditional text classification tasks take sentiment or topic as the main output label, the SuperGLUE \footnote{https://super.gluebenchmark.com/} benchmark has witnessed over 90\% accuracy for most of the text classification tasks. The question is: for real world text classification tasks, can ChatGPT achieve over 90\% accuracy? We reviewed 10 text classification tasks and the answer is No. 

A few works have showed the appealing classification performance of ChatGPT in the zero shot setting. For example, \citep{heck-etal-2023-chatgpt} evaluate ChatGPT on dialogue state tracking and show that it achieved an average accuracy of 56.44\%, which was the state-of-art in the zero shot setting but still cannot match for supervised systems.  \citep{zhao-etal-2023-hw} explore the feasibility of using ChatGPT and prompt learning for text entailment classification and show that it is competitive with BERT based zero shot models. Other pieces of works argue that GhatGPT still cannot beat with fine-tuned Transformer or BERT models. For example, \citep{ghanadian-etal-2023-chatgpt} conduct evaluation for ChatGPT on suicide risk assessment and show that zero-shot ChatGPT reached 0.73 accuracy while the fine-tuned ALBERT achieved 0.86 accuracy, also they find the few-shot ChatGPT was even not as good as the zero-shot one. \citep{kim-etal-2023-chatgpt} also show that ChatGPT was still behind the existing fine-tuned BERT models by a large margin on the task of science claim classification. 

More recently, \citep{koopman2023dr} evaluate ChatGPT on 100 topics from the the TREC 2021 and 2022 Health Misinformation track, the accuracy of ChatGPT dropped from 80\% to 33\% when prompting for "Yes/No" and "Yes/No/Unsure" answers, and it further dropped to less than 60\% when the prompts are paraphrased under the same meaning. \citep{song-etal-2023-large} explore ChatGPT on general in-domain (GID) and out-of-domain intent discovery and recognition, and found that for in-domain the overall performance of ChatGPT is inferior to that of the fine-tuned baselines and for out-of-domain intent discovery ChatGPT performs far worse than the fine-tuned baselines under multi-sample or multi-category scenarios, it was also noted ChatGPT cannot conduct knowledge transfer from in-domain demonstrations and generalize it to out-of-domain tasks. \citet{tan-etal-2023-chatgpt} evaluate both ChatGPT and GPT4 on conversation sentiment classification, both systems achieved comparable performance (between 40\%-60\% accuracy) to the supervised models on two out of three data sets. 

The other promising work is LLM-generated text detection with ChatGPT, \citep{zhu2023beat} develop a pipeline experimented with six datasets and the average accuracy reached 90.05\%, compared to other zero shot method which achieved to 60\%-70\% only, but this is a fairly easy binary classification task the pipeline engineering tricks may not available for other text classification tasks.    

There are several key findings for the classification tasks: First, ChatGPT's performance is good under the zero shot classification setting but still fall behind the supervised models. Second, when the label space increases, the accuracy of ChatGPT drops significantly. Third, knowledge transfer can hardly happen for out-of-distribution classification tasks even though some demonstrations are given. Fourth, ChatGPT's classification performance tends to be good when there is public data which is related to the target task. 

\begin{table*}[!tb]
    \centering
    \small
\begin{tabular}{ lllc} 
 \toprule
 Tasks & Example Application & Related Works & Beat state-of-art \\
 \midrule
 \multirow{6}{*}{Classification} 
    & Health misinformation detection & \citep{koopman2023dr} & 0 \\
    \cmidrule{2-4}
    & Intent discovery and recognition &  \citep{song-etal-2023-large} & 0 \\
    \cmidrule{2-4}
    & Conversation sentiment analysis & \citep{tan-etal-2023-chatgpt} & 1 \\
    \cmidrule{2-4}
    & Text entailment & \citep{zhao-etal-2023-hw} & 0 \\
    \cmidrule{2-4}
    & LLM generated text detection &  \citep{zhu2023beat} & 1 \\
    \hline
 
  \multirow{2}{*}{Generation} 
    &Summarization  &  \citep{bang2023multitask,caciularu-etal-2023-peek} & 0 \\
    \cmidrule{2-4}
    & Question Answering &   \citep{bai2023benchmarking,wang2023evaluating}  &  1\\
    \cmidrule{2-4}
    & Machine Translation   &   \citep{bang2023multitask,jiao2023chatgpt} & 0.5 \\
    \cmidrule{2-4}
    & Paraphrasing  &   \citep{cegin-etal-2023-chatgpt} & 1 \\
    \cmidrule{2-4}
    & Controllable Generation  &   \citep{pu-demberg-2023-chatgpt,valentini-etal-2023-automatic} & 0 \\
    \hline
    Sequence labelling & Named entity recognition & \citep{xie2023empirical} & 0  \\
   \hline
   \multirow{2}{*}{Information Retrieval} 
    & Query rewriting  &  \citep{wang-etal-2023-query2doc,mao-etal-2023-large} & 1 \\
    \cmidrule{2-4}
    & Passage retrieval &   \citep{muennighoff2022sgpt,sun2023learning} & 1 \\
    \cmidrule{2-4}
    & Passage re-ranking  &   \citep{sachan2022improving} & 1 \\
    \cmidrule{2-4}
    & Content re-generation  &   \citep{yu2022generate} & 1 \\
    \hline
    Parsing & Text-to-SQL parsing & \citep{sun2023battle} & 0  \\
    \hline
    \multirow{2}{*}{Reasoning} 
    & Arguments on reasoning  &  \citep{huang2022towards,huang2023large}  & 0 \\
    \cmidrule{2-4}
    & Reasoning skills  &   \citep{qin2023chatgpt}  & 0.5 \\
    
    \hline
    \multirow{2}{*}{Multilingualism} 
    & Back Translation  &  \citep{zhang2023don} & 0 \\
    \cmidrule{2-4}
    & Arabic Language Analysis &   \citep{khondaker2023gptaraeval} & 0 \\
 \bottomrule
\end{tabular}
\caption{ChatGPT for various NLP tasks: 0/0.5/1 refer to the evaluation results that ChatGPT's performance is worse/comparative/better than the state-of-art methods. It can be seen that ChatGPT is well adapted to text generation and information retrieval, but not directly suitable for other NLP tasks.  }
\label{tab:nlp_tasks_chatGPT}
\end{table*}

\begin{table*}[!tb]
    \centering
    \small
\begin{tabular}{ llp{10cm} } 
 \toprule
 Perspective & Example & Related Works \\
 \midrule
 \multirow{4}{*}{Social implications} 
    & Bias &   \citep{ray2023chatgpt,zhang2023don,wang-etal-2023-primacy} \\
    \cmidrule{2-3}
    & Fairness &   \citep{deshpande-etal-2023-toxicity} \\
    \cmidrule{2-3}
    & Ethics & \citep{stahl2024ethics} \\
    \cmidrule{2-3}
    & Employment & Impact on jobs \citep{george2023chatgpt} \\
     \cmidrule{2-3}
    & Energy &  Renewable and Sustainable Energy \citep{rane2023contribution} \\
    \hline
 
  \multirow{3}{*}{Safety} 
    & Privacy  &   \citep{kassem-etal-2023-preserving,wu-etal-2023-depn} \\
    \cmidrule{2-3}
    & Misinformation &  \citep{li2023does} \\
    \cmidrule{2-3}
    & Attach and Defence  &  \citep{zou2023universal} \\
 \bottomrule
\end{tabular}
\caption{Social implications and safety of ChatGPT.}
\label{tab:social_safety}
\end{table*}

\subsection{Generation}
\subsubsection{Summarization}
Text summarization aims to convert text or collection of text into short text containing key information. Conciseness is one of the main goals of summarization, while some works have pointed out that summaries generated by ChatGPT tend to be redundant when there are no length constraints in prompts. This can be improved by using restricted prompts, thus achieving a balance between precision and recall.

Although ChatGPT performs well under zero-shot settings, it still under-performs fine-tuned state-of-the-art models based on automatic evaluation metrics. For example, fine-tuned BART outperforms zero-shot ChatGPT by a large margin. In multi-document summarization, \citep{caciularu-etal-2023-peek} design novel pre-training objective and the their model significantly outperforms GPT-based large language models. As for specific domains like biomedical tasks, ChatGPT performs much worse than fine-tuned BioBART in datasets that have dedicated training sets. Nevertheless, when lacking large training data, zero-shot ChatGPT is more useful than domain-specific fine-tuned models, showing its good zero-shot performance.

Summaries generated by ChatGPT tend to be preferred by humans, for they have fewer grammatical errors and are more fluent and coherent. At the same time, informativeness of these summaries is not high. The use of Reinforcement Learning from Human Feedback leads ChatGPT to have a tendency to focus on linguistic aspects but struggle to ensure fidelity to the factual information and alignment with the original source, so it may overfit unconstrained human evaluation, which is affected by annotators’ prior, input-agnostic preferences.

Sometimes ChatGPT is not trustworthy, especially in specialized and technical domains, as it may confidently produce factually incorrect outputs. \citep{ye-etal-2023-cp} have confirmed that on binary code summarization, ChatGPT only have a rudimentary understanding of assembly code, without any higher-level abstract semantic comprehension. In contract summarization, \citep{sancheti-etal-2023-read} have pointed out that hallucinations in ChatGPT make it hard to perform this task. Some works have come to opposite conclusions, because prompt design can greatly affect ChatGPT's performance. For example, \citep{qin-etal-2023-chatgpt} find that controlling the length of summaries may harm ChatGPT’s summarization ability, which is contrary to contents mentioned earlier in this section, indicating the instability of ChatGPT.

In summary, ChatGPT performs well on the zero-shot summarization, its summaries are more in line with human preferences, but it under-performs fine-tuned models based on automatic evaluation metrics, and informativeness is not high. Besides, ChatGPT is not likely to play a role in specialized-domain summarization due to hallucinations and instability.

\subsubsection{Question Answering and Dialogue}
Question answering (QA) and dialogue tasks can assess the retrieval, understanding, and generation capabilities of ChatGPT. In open-domain question answering, \citep{bai2023benchmarking} have benchmarked several large language models and proved ChatGPT's near-perfect performance. When providing false presupposition, ChatGPT performs well in explicitly pointing out the false presupposition. In practice, ChatGPT performs comparably to traditional retrieval-based methods, but falls behind newer language models like Bing Chat. \citep{nov2023putting} use ChatGPT to answer health questions, responses toward patients’ trust in chatbots’ functions are weakly positive, and laypeople appear to trust the use of chatbots to answer lower risk health questions. But as the complexity of the QA task increases, people's trust in ChatGPT responses decreases. Besides, ChatGPT does not perform well on low resource QA, regardless of the use of broad and diversified training corpora.

\citep{feng-etal-2023-towards} have confirmed ChatGPT's superiority over previous methods in dialogue state tracking, while smaller fine-tuned model can achieve comparable performance. In open-domain dialogue, ChatGPT can generate fluent response, while lags behind fine-tuned GPT-2 on automatic evaluation metrics and slightly under-performs Claude in certain configurations. In task-oriented dialogue, ChatGPT often struggles to differentiate subtle differences among the retrieved knowledge base records when generating responses, and it tends to generate hallucinated information beyond the given knowledge.

In summary, ChatGPT performs well in simple open-domain tasks, where humans prefer its responses. While in complex, low resource or task-oriented scenarios, it still has a lot of room for improvement.

\subsubsection{Machine Translation}
Machine translation refers to the conversion of a kind of natural source language into another target language. As one of the most common usages of large language models, it greatly facilitates modern life. ChatGPT translates well between high-resource languages (such as European languages), even on par with commercial systems, but when faced with low-resource languages, it lags behind fine-tuned models as well as commercial systems. Another common finding is that ChatGPT translates well in XX → Eng tasks, but it still lacks the ability to perform Eng → XX translation. Similar to other tasks, ChatGPT is unstable on machine translation, it sometimes exhibits omissions and obvious copying behaviors.

One thing that makes ChatGPT different from other translation systems is that ChatGPT can better model long-term dependencies and capture discourse-level information, while other systems focus more on word-level accuracy, resulting in ChatGPT being more preferred by humans. In addition, ChatGPT has the ability for zero pronoun resolution and recovery, which is one of the most difficult problems in NLP. Therefore, we can draw a conclusion that the translations generated by ChatGPT are more focused on overall linguistic quality and perform extremely well on high-resource languages, while for word-level accuracy and low-resource languages, ChatGPT does not perform as well as fine-tuned models.

\subsubsection{Paraphrasing and Data Augmentation}
ChatGPT is efficient and cost-effective for tasks such as data augmentation and paraphrasing. Several work have shown that ChatGPT can produce data with higher diversity in paraphrase generation and show similar model robustness to the data collected from human workers. For example, \citep{jon-bojar-2023-breeding} use ChatGPT to generate 40 diverse paraphrases for one sentence, and \citep{michail2023uzh_clyp} use synthetic tweets generated by ChatGPT as training data, these methods have achieved desired effect.

But ChatGPT also has drawbacks, it does not produce alternative names for named entities, e.g. locations, songs, people, which is something crowd-sourced data handles well. Nevertheless, it is still an efficient and cost-effective option for these tasks.

\subsubsection{Controllable Generation}
Controllable generation aims to generate text with specific characteristics, which has attracted much attention in recent years. Although ChatGPT can fit human preferences, it does not perform well on controllable generation tasks. \citep{pu-demberg-2023-chatgpt} prompt ChatGPT to generate different summaries for laymen and experts, although it outperforms previous state-of-the-art models, the generated summaries are quite different from human-written texts. Some works have found that ChatGPT does not to follow numerical restrictions properly, which may be caused by incorrect tokenization. Another research by \citep{valentini-etal-2023-automatic} try to prompt ChatGPT to generate stories for children of different ages, they find that ChatGPT cannot avoid using complex words so the generated stories are significantly less readable than human-written stories. Besides, ChatGPT performs poorly on some other tasks, such as decontextualization and complex controlled paraphrase generation.

Although ChatGPT struggles at the above fine-grained hard constraints, it can deal with coarse constraints. For example, zero-shot ChatGPT outperforms supervised baselines on content-constrained text generation such as sentiment and keyword constraints. In addition, it can continue writing more fluent and coherent stories given the beginning text of a story. ChatGPT is good at mimicking rather than mastering complex understanding, organization and generation abilities, so it does not cope well with hard control signals.

\subsubsection{Other Generation Tasks}
ChatGPT can be used in various generation tasks. For code generation, \citep{liu2023your} show that two open-source models can out-perform ChatGPT according to their proposed evaluation framework. \citep{singh-etal-2023-codefusion} point out that ChatGPT under-performs T5 on code generation for Bash and CF rules. \citep{xiao-etal-2023-evaluating} utilize ChatGPT in the field of education, the reading passages and corresponding exercise questions which are all generated by ChatGPT are suitable for students and even surpass the quality of existing human-written ones. There are other generation tasks using ChatGPT, such as generating explanations for multiple-choice item options in reading comprehension tests, providing segments and prompts to derive question-answering pairs, giving a sentence and an entity to generate a related question for this entity, ChatGPT performs well on these tasks, and in some cases it is even comparable to humans.

When it comes to generating constructive or complex contents, ChatGPT tends not to perform very well. For example, in education ChatGPT can act as a teaching coach, but it still has room for improvement in generating insightful and novel feedback for teaching. \citep{jentzsch-kersting-2023-chatgpt} try to figure out whether ChatGPT has sense of humor. ChaGPT can only identify, reproduce, and explain puns that fit into the fixed pattern, and it cannot produce original funny contents, therefore, ChatGPT just has the ability to learn specific joke patterns instead of being really funny.

For text generation tasks, the main advantage of ChatGPT is that it has good writing ability and alignment to human values, which explains why it performs on par with humans on some simple generation tasks. But it still does not perform well on tasks that require generating creative contents or understanding complex semantic features and syntactic parses.

\subsection{Sequence Labelling}
Sequence labelling assigns tags to words or phrases in a sequential manner, such as Named Entity Recognition (NER), Part-of-Speech tagging and noun phrase recognition. Traditional statistical approaches use Hidden Markov Model (HMM) or Conditional Random Fields (CRF) for sequence labelling, the current state-of-art methods rely on deep representation with CRF. Different from classification or generation tasks, the labels for sequence labelling tasks are towards local text spans, and the feature representation for the text spans is often limited to a small window. \citep{xie2023empirical} conduct an empirical study for zero-shot NER with ChatGPT, in which the authors explore a decomposed question-answering paradigm by breaking down the NER task into simpler sub-problems by labels. They also experiment with syntactic prompting and tool augmentation, verifying the effectiveness of their methods on Chinese and English Scenarios, and on both domain-specific and general-domain datasets. 

\subsection{Information Retrieval}
Typical information retrieval systems contain two steps: Given a query, search relevant documents in the first step and rank the returned documents in the second step. Many works have applied LLMs in to the information retrieval process. There are four general modules that LLMs can assist: rewriter, retriever, reranker and reader. Rewriter is an essential IR module that aims to improve the precision and expressiveness of user queries. Query rewriting acts in two circumstances: ad-hoc retrieval, which bridges the vocabulary mismatch between query and documents, and conversational search, which iteratively refines and adapts system responses based on evolving conversations. For example, Query2Doc \citep{wang-etal-2023-query2doc} generates pseudo-documents by few-shot prompting LLMs, and then expands the query with generated pseudo documents.  LLM4CS \citep{mao-etal-2023-large} leverages ChatGPT as a text-based search intent interpreter to help conversational search, three prompting methods were used to generate multiple query rewrites and hypothetical responses, these query rewrites were aggregated into an integrated user query representation. 
The retriever is typically used in the early stage of IR for recall improvement, the classical bag-of-words model BM25 has demonstrated strong robustness in many retrieval tasks. SGPT \citep{muennighoff2022sgpt} modifies GPT models into cross or bi-encoders for semantic search. GENRET \citep{sun2023learning} learns to tokenize documents into short discrete representations via a discrete auto-encoding approach.  
The re-ranker is another important module that returns an ordered list of relevant documents.  It serves as key part for fine-grained document filtering. UPR \citep{sachan2022improving} scores the retrieved passages with LLMs and re-ranks the passages based on the log-likelihood score over the question. The reader or content generation can be regarded as the final process of information management, which can compress the searched text into a user friendly output. GenRead \citep{yu2022generate} first prompts a large language model to generate contextual documents based on a given question, and then reads the generated documents to produce the final answer.     

Although the generative ability brings some discrepancy between the pre-training objectives of LLMs and the ranking objectives, two recent studies \citep{zhang2023preliminary} have shown that ChatGPT reached competitive results on the task of IR compared with strong baselines. \citep{zhang2023preliminary} empirically evaluated ChatGPT on retrieving requirements information from specialized and general artifacts, under the zero-shot setting both quantitative and qualitative results revealed ChatGPT's promising ability to retrieve requirements relevant information (high recall) and limited ability to retrieve more specific requirements information (low precision). Similarly, \citep{sun2023chatgpt} explored instructional methods for ChatGPT on various passage re-ranking benchmarks and validated its capability to supervised models. 

\subsection{Parsing}
Parsing was the backbone in the statistical age of NLP.  Many high level NLP tasks, e.g., Machine Translation and Information Extraction, rely on constituency parsing or dependency parsing. In the neural age, the parsing step can be skipped as the distributed representation of the text can be directly fed to the downstream tasks. Therefore, nowadays parsing tasks can often be solved by sequence-to-sequence learning. \citep{sun2023battle} compared ChatGPT with other five open source LLMs on the text-to-SQL parsing task.  It is found that  the open-source models fall significantly short in performance when compared to closed-source models.However, it is worth noting that even GPT-3.5 performs worse than smaller baseline models on several classical text-to-SQL parsing datasets. 

\subsection{Reasoning}
Reasoning is a fundamental part of human intelligence, it is the process of thinking about a premise in a logical and systematic way based on past experience or context. It is still not clear whether LLMs have real reasoning abilities. Starting from the early claim that "LLMs are few shot learners", other similar claims in terms of reasoning include "LLMs are decent zero-shot reasoners", and "LLMs are still far from achieving acceptable performance on common planning/reasoning tasks",  or LLMs cannot self-correct reasoning yet.  After overviewing the techniques for improving reasoning skills in LLMs as well as the methods and benchmarks for evaluating reasoning abilities, we find that it is not clear whether LLMs are marking predictions based on true reasoning or heuristics. Their ability to reason step-by-step and return rationales may be incorrect and inconsistent. 

A few recent work analyzed the specific reasoning abilities of ChatGPT. For example, \citep{jang-lukasiewicz-2023-consistency} investigated four properties in logical reasoning: semantic equivalence, negation, symmetric property and transition, and showed that ChatGPT exhibited enhanced negation and transitive consistency but still made mistakes that violate the logical properties. Also, ChatGPT frequently changes its answer when the input text is paraphrased or when the order of the input sentences is switched. \citep{wang2023can} explored ChatGPT's reasoning ability via debate: they first obtained ChatGPT’s initial solution and performed evaluation on examples where it achieved a correct answer. Then they synthesized an invalid solution abductively by conditioning on a wrong target answer. Afterward, they initiated a debate-like dialogue between ChatGPT and the user (simulated by ChatGPT conditioned on the invalid solution), in order to see whether ChatGPT can hold and defend its belief in truth during the debate. It is found that ChatGPT's belief and disbelief is not robust and could be easily perturbed by the user and it frequently admits to or is misled by users' invalid responses/arguments, despite being able to generate correct solutions in the beginning. \citep{qin2023chatgpt} experimented with ChatGPT with arithmetic, commonsense, symbolic and logical reasoning, and showed that ChatGPT achieves better performance than GPT-3.5 when using chain-of-thought in arithmetic reasoning, but using chain-of -thought does not always provide better performance in commonsense reasoning, and it even performs worse than GPT-3.5 in many cases for symbolic and logical reasoning. In a legal application, \citep{kang-etal-2023-chatgpt} showed that ChatGPT can produce reasonable answers but mostly fail to yield correct reasoning paths aligned with legal experts. Its performance can be improved by providing parts of the annotated reasoning paths, including similar annotated scenarios for in-context learning and decomposing complex issues into simpler questions. 

In summary, we find that it is not clear whether ChatGPT has real reasoning skills or make the predictions just based on the memory. Also, techniques like chain-of-thought would result in inconsistent generations. Prompting and using larger language models may not be the final solution for solving reasoning problems. We suggest to incorporate heuristics and probability (e.g. Bayes Networks) inference in LLMs to enhance their reasoning abilities. 

\subsection{Multilingualism}
Many studies have shown that ChatGPT's performance is better when the input language is English, a big reason is that the training datasets are heavily skewed towards English. \citep{zhang2023don} employed a prompt back-translation approach and showed that ChatGPT can return consistent results in translation-equivalent tasks but struggle to provide accurate answers in translation-variant tasks. \citep{khondaker2023gptaraeval} evaluated ChatGPT on 44 Arabic language understanding and generation tasks and found that it can be consistently surpassed by smaller models that have been fine tuned on Arabic. 

\subsection{Mixed Tasks}
There are a few studies which investigated whether ChatGPT is a general purpose solver for a specific domain. \citep{jahan2023evaluation} showed ChatGPT performs quite poorly in comparison to the fine-tuned models (BioGPT and BioBART) in the biomedical domain, whereas it outperforms fine-tuned models on datasets where the training data size is small. \citep{li2023chatgpt} conducted an empirical study of ChatGPT and indicate it can compete with fine tuned models in the financial domain, but still fall behind on tasks when deeper semantics and structural analysis are needed.  

\section{Social implications and Safety} We list the recent works on social implications and safety concerns of ChatGPT in Table 2.
\paragraph{Social implications} Bias and fairness arise from the philosophical concept that humans  should be treated equally by models. However, bias may be caused by unintentional behaviours and come from different sources, including training data collection, model design and human interaction and annotation. Recent studies \citep{ray2023chatgpt}  have classified bias into different categories, such as racial and gender bias, language bias, culture and linguistic bias, geographic bias, etc. ChatGPT has shown strong language bias towards English, \citep{wang-etal-2023-primacy} investigated the primacy effect of ChatGPT and showed that  ChatGPT’s decision is sensitive to the order of labels in the prompt and it has a clearly higher chance to select the labels at earlier positions as the answer. \citep{espana-bonet-2023-multilingual} found the political and language bias of ChatGPT in 2023: Between Feb and Aug 2023, ChatGPT transitioned from a left to neutral political orientation, with a right-leaning period in the middle for English and Spanish, while its current version as of Aug 2023 consistently shows Left-leaning for 4 languages. \citep{deshpande-etal-2023-toxicity} revealed that ChatGPT when assigned a persona can be significantly toxic and unsafe when compared to the default setting. Other social factors also consider ethics \citep{stahl2024ethics}, impact on employment \citep{george2023chatgpt} and consumption of energy \citep{rane2023contribution}. 
\paragraph{Safety}  Privacy attacks on ChatGPT are implemented by constant cue modification. Privacy of training data in ChatGPT is extracted by multiple rounds of prompt modification. A few studies investigated privacy preservation for open source LLMs, which mainly follow two approaches: unlearning with modifying the learning objective function, or editing selected neurons directly. \citep{huang-etal-2023-privacy} found that kNN-LMs are more susceptible to leaking private information from their private datastore than parametric models for retrieval-based language models. Another safety concern is the the spread of misinformation, \citep{li2023does} showed that ChatGPT can poison data and mislead fake news detection systems trained using real-life news. Meanwhile, adversarial attacks on ChatGPT can be done by appending a special sequence of characters to a user query, which will cause the system to obey user commands even if it produces harmful content.

\section{Performance Over Time}
Machine Learning model generalisation ability is vital for applications on unseen data. The claim that large language models are few shot learners is argued by many researchers as some LLMs have seen task examples during pre-training for a range of tasks, and are therefore no longer zero or few-shot for these tasks. Additionally, for classification tasks with no possibility of task contamination, \citep{li2023task} showed that LLMs rarely demonstrate statistically significant improvements over simple majority baselines, in both zero and few-shot settings. \citep{chen2023chatgpt} also demonstrated that the behavior of GPT-3.5 and GPT-4 has varied significantly over a relatively short amount of time, for example, GPT-4 were less willing to answer sensitive questions in June than in March 2023, and both GPT-4 and GPT-3.5 had more formatting mistakes in code generation in June than in March 2023.

\section{Challenges and Opportunities}
It is also noted that recent LLM leaderboards, e.g., AlpacaEval\footnote{\url{https://tatsu-lab.github.io/alpaca_eval/}} and Chatbot Arena Leaderboard\footnote{\url{https://huggingface.co/spaces/lmsys/chatbot-arena-leaderboard}}, display that some open source models with less parameters have achieved similar or better NLP ability to gpt-3.5-turbo. However, it is unclear whether there is test data contamination for the public models. We identify three key challenges for both closed and open large language model evaluation:
\paragraph{Explainability} Providing meaningful explanations is a key part of a trustworthy system. ChatGPT can provide answers with explanations. However, some studies have shown that the explanations returned by ChatGPT are not consistent within the context. We hereby illustrate two ways for explainable LLMs: one is to build multi-agent systems and allocate specific agents for the explanability function, the other way is to equip the instruction training data with explainable items which could be more expensive. 
\paragraph{Continual Learning} As new data and tasks arrives, the ability of continual learning is important. Typical continual learning approaches, such as memory replay, regularization and model architecture re-design, are not scalable for LLMs like ChatGPT. More efficient ways use frozen and fine tune approaches, such as Adapter \citep{pfeiffer2020adapterhub} and LoRA \citep{hu2021lora}. The recent mixture-of-experts approaches \citep{diao2023mixture} facilitate a trade-off between learning and forgetting and can be a more applied way for LLM continual learning.
\paragraph{Lightweight Modeling} The big size of ChatGPT limits itself to local deployment. Recent works study small language modelling, including model distillation from LLM, training small sized LM on larger data sets, and over-parameterization. However, it is still not clear whether larger data on small model or small data on larger pre-trained model performs better for generative language modelling.
\section{Conclusion}
The research on large language modelling is ongoing and reliable model evaluation is important. In this paper, we review ChatGPT's real performance levels on different NLP tasks and find that it is often under-performed to fine tuned models in many tasks. The zero or few shot learning ability is largely dependent on its large training data, which is not open to the public. The performance of its ability is degrading over time, which may hinder its widespread applications.  Moreover, its bias is significant and varies over time.

\section*{Acknowledgements}
Ming Liu is supported by the Australia Research Council Linkage Project (LP220200746).

\section*{Author Contributions}
M.L. designed the content structure, M.L. and R.L. wrote the main manuscript text, Y.Z. and Y.Q wrote the parsing section, R.L. wrote the sequence labeling section, Y.P. wrote the reasoning section, H.W. and B.W. supervised this project. All authors reviewed the manuscript.

\bibliography{ref}

\end{document}